\documentclass[a4paper]{article}
\usepackage[margin=1in]{geometry}

\usepackage[utf8]{inputenc} % allow utf-8 input
\usepackage[T1]{fontenc}
\usepackage{lmodern}
\usepackage{hyperref}       % hyperlinks
\usepackage{url}            % simple URL typesetting
\usepackage{booktabs}       % professional-quality tables
\usepackage{amsfonts}       % blackboard math symbols
\usepackage{nicefrac}       % compact symbols for 1/2, etc.
\usepackage{microtype}      % microtypography
\usepackage{xcolor}         % colors

\usepackage{natbib}

\usepackage{amsmath}
\usepackage{amsthm}
\usepackage{amsfonts}
\usepackage{amssymb}

\usepackage{subcaption}

\usepackage[ruled,linesnumbered]{algorithm2e}
\usepackage{algorithmicx}
\usepackage{algpseudocode}
\usepackage{bbm}
\usepackage{booktabs}       % professional-quality tables
\usepackage[T1]{fontenc}    % use 8-bit T1 fonts
\usepackage{graphicx}
\usepackage{hyperref}       % hyperlinks
\usepackage{mathrsfs}
\usepackage{microtype}      % microtypography
\usepackage{multirow} % for cmd 'multirow', 'multicolumn'
\usepackage{nicefrac}       % compact symbols for 1/2, etc.
\usepackage{nccmath}
\usepackage{rotating}
\usepackage{
subcaption} % for 'subtable' env
\usepackage{tikz}
\usepackage{url}            % simple URL typesetting
\usepackage[utf8]{inputenc} % allow utf-8 input
\usepackage{xcolor}         % colors
\usetikzlibrary{positioning, calc, shapes.geometric, shapes.multipart, 
	shapes, arrows.meta, arrows, 
	decorations.markings, external, trees}
\tikzstyle{ann} = [draw=none,fill=none,right]

\usepackage{pifont}
\newcommand{\no}{\ding{55}}
\newcommand{\yes}{\checkmark}

\newcommand{\mat}[1]{\mathbf{#1}}
\renewcommand{\vec}[1]{#1}

\usepackage{tikz}
\usepackage{pgf}
\usetikzlibrary{shadows,arrows,calc}
\usetikzlibrary{positioning,shapes.geometric}
\usetikzlibrary {shapes.misc} 
\usetikzlibrary{intersections}

\title{Causal Learning in Biomedical Applications: \\
Krebs Cycle as a Benchmark}

\begin{document}

\author{%
  Xiaoyu He \\
  AI Center \\
   Czech Technical University in Prague \\
   Karlovo namesti 13, Prague 12135  \\
   \\
  Petr Ryšavý 
  \\
  Department of Computer Science \\
  Czech Technical University in Prague \\
   Karlovo namesti 13, Prague 12135  \\
  \texttt{petr.rysavy@fel.cvut.cz} \\
  % examples of more authors
  \\
  Jakub Mare{\v c}ek \\
  AI Center \\
   Czech Technical University in Prague \\
   Karlovo namesti 13, Prague 12135  \\
  \texttt{jakub.marecek@fel.cvut.cz} \\
}

\maketitle

\begin{abstract}
\textbf{Background:} Learning causal relationships from time series data is an important but challenging problem. Existing synthetic datasets often contain hidden artifacts that can be exploited by causal discovery methods, reducing their usefulness for benchmarking. 

\textbf{Methods:} We present a new benchmark dataset based on simulations of the Krebs cycle, a key biochemical pathway. The data are generated using a particle-based simulator that models molecular interactions in a controlled environment. Four distinct scenarios are provided, varying in time series length, number of samples, and intervention settings. 

\textbf{Results:} The benchmark includes ground-truth causal graphs for evaluation. It supports quantitative comparisons using metrics such as Structural Hamming Distance, Structural Intervention Distance, and F1-score. A comprehensive evaluation of 14 causal discovery methods from different modelling paradigms is presented. Performance is compared across datasets using multiple accuracy and efficiency measures. 

\textbf{Conclusions:} The dataset provides a reproducible, interpretable, and non-trivial benchmark for testing causal learning methods on time series data. It avoids common pitfalls such as residual structural patterns and supports interventions and evaluation with known causal ground truth. This makes it a useful tool for the development and comparison of causal discovery algorithms.\end{abstract}

\section{Introduction}

Understanding causal models is important in a number of fields, from healthcare to economics, as it allows for precise forecasting and training of reinforcement learning algorithms. Learning causal models involves extracting potential non-linear relationships and dependencies between variables from sampled time series. For example, the modelling of biomarkers of non-communicable disease as a function of diet and action monitoring has shown the potential of being a powerful tool to guide the recommendations for a healthy diet. 

The causal learning community agrees that there is a need for better synthetic datasets to test causal learning algorithms\citep{reisach1, 
standardizing2, reisach3}. Many synthetic dataset benchmarks suffer from residual information in the data that the $R^2$-sortability can identify. In the case of real-world datasets, we often cannot be sure what the ground-truth causal relationships are. Often, datasets for causal discovery are too large, and as a result, they are sampled without any standardized sampling approach, thus making different papers using the datasets incomparable.

In this paper, we aim to fill this gap and provide a standardized synthetic dataset that does not suffer from the problems mentioned above. The dataset is based on simulating a set of chemical reactions describing the Krebs cycle, and for that, it uses a publicly available generator at \cite{github-nagro}. The randomness in the data is caused by simulating the molecules in a box and providing the molecules with locations and velocities. Whenever molecules forming the left-hand side of a reaction meet, they are replaced with reactants as given by the equation.

First, we provide a brief review of a variety of methods that can be used in causal learning. Later, we provide a list of requirements that we can expect from such causal learning methods to illustrate their expressiveness. A discussion of which criteria are supported by the existing methods follows. Section \ref{sec:benchmark} explains the dataset in detail and shows a possible evaluation of methods on the dataset. We show a comparison on 4 datasets. Next, we compare the presented dataset with other causal learning datasets. In conclusion, we give preference to public repositories where the dataset, as well as the source code for the evaluation of the method, can be found.

\section{Preliminaries and Related Work}

Learning most causal models involves solving NP-hard non-convex optimization problems. Just as there is ``one'' convex optimization and ``many'' non-convex optimization problems, there are many causal models and methods for learning them. Perhaps the most elegant approach to causal learning utili'ses techniques from system identification. 

\paragraph{System Identification and Linear Dynamic Systems (LDS)}
\label{sec:systemid}
Let $m$ be the hidden state dimension and $n$ be the observational dimension. 
%A linear dynamic system (LDS) $\mathbf{L}$ is defined as a quadruple $ (\mathbf{F}, \mathbf{G}, \mathbf{\Sigma}, \mathbf{V})$, where $\mathbf{F}$ and $\mathbf{G}$ are \emph{system matrices} of dimension $m \times m$ and $n \times m$, respectively. $\mathbf{\Sigma} \in \mathbb{R}^{m \times m}$ and $\mathbf{V} \in \mathbb{R}^{n \times n}$ are covariance matrices \cite{west2006bayesian}. 
A single realisation of the LDS of length $T$, denoted $\mathbf{X} = \{x_1, x_2, \ldots, x_T\} \in \mathbb{R}^{n \times s \times T}$ is defined by \emph{initial conditions} $\vec{\phi}_0$ and \emph{system matrices} $\mathbf{F}$ and $\mathbf{G}$ as 
\begin{align}
    \phi_t &= \mat{F}\phi_{t-1} + \omega_t, 
    \label{eq:lds1} &&\omega_{t}\sim N(0, \Sigma) \in \mathbb{R}^{m \times s}\\
    x_t &= \mat{G}' \phi_t  + \upsilon_t && \upsilon_{t}\sim N(0, \mat{V})\in \mathbb{R}^{n \times s},
    \label{eq:lds2}
\end{align}
where $\phi_t \in \mathbb{R}^{m \times s}$ is the vector autoregressive processes with hidden components and $\left\{\omega_t,\upsilon_t\right\}_{t \in \{1, 2, \ldots, T\}}$ are normally distributed process and observation noises with zero mean and covariance of $\Sigma$ and $\mat{V}$ respectively.
The transpose of $\mat{G}$ is denoted as $\mat{G}'$. Vector $x_t \in \mathbb{R}^{n \times s}$ is the observed output of the system. 
In non-linear dynamical systems, one replaces the multiplication $\mat{F}\phi_{t-1}$ with a function $f(\phi_{t-1})$.
It is well known that there are multiple, equivalent conditions for the identifiability of $\mat{F}, \mat{G}$, given by so-called Hankel matrices, conditions on the transfer function, or frequency-domain conditions, among others\citep{willems2005note}.  
The is also a recent understanding of sample complexity of the problem\citep{tsiamis2023statistical}.

%Recently, Zhou and Marecek \cite{zhou2020proper} proposed to find the global optimum of the objective function subject to the feasibility constraints arising from \eqref{eq:lds1} and \eqref{eq:lds2}:
%\begin{equation}\min_{f_t,\phi_t,\mat{G},\mat{F},\omega_t,\upsilon_t} \sum_{t \in \{1,2, \ldots, T\}} \|X_t - f_t \|_2^2 + \|\omega_t\|_2^2 + \|\upsilon_t\|_2^2,\label{eq:optcriterion}
%\end{equation}
%for a $L_2$-norm $\|\cdot\|_2$. In the causality problem we are given $N$ variables $\mat{X_j} \in \mathbb{R}^{s \times T}$. A natural problem is to find the estimation values $f_t$ of the LDS that generated the observation data. In other words, we are interested in finding the optimal objective values and the residual vectors $\upsilon_t$, $\omega_t$ that belong to each LDS.

\paragraph{Linear Additive Noise Models} 

%\textcolor{red}{Explain assumptions of linearity }
%\textcolor{red}{Jakub to paste in some tables on the computational complexity}

Throughout causal modelling, one wishes to learn a function $f$, which is known as the structural assignment map and is closely related to the $f$ above.
Under the assumption that the structural assignments are linear, noises $N_j,j=1,\dots,N$ are independently identically distributed (i.i.d.) and follow the same Gaussian distribution, or alternatively, noises $N_j,j=1,\dots,N$ are jointly independent, non-Gaussian with strictly positive density, 
one obtains linear additive noise models (ANM).
In studying ANM, one may benefit from a long tradition of work on linear system identification. 
In particular, the identifiability of linear ANM can be reduced to the identifiability of linear dynamical systems (cf. Proposition~7.5 \& Theorem~7.6 in \cite{peters2017elements}).
%(cf. Table \ref{ID_TAB:Sys_ID_Fully} for fully-observed systems and Table~\ref{ID_TAB:Sys_ID_Fully2} more broadly). 

%In this section, we will focus on some common approaches to causal learning. The field itself is vast; Google Scholar offers more than $4.5$ millions of hits for the term \emph{causal learning}, only in the past year of 2023, the number of hits is almost $100{,}000$ scientific papers. As a result, this brief review can never be complete.

\paragraph{Bayesian networks} Another classic example in causal learning are \emph{Bayesian networks}, first introduced by Pearl in 1985\citep{pearl1985bayesian}. Bayesian networks are formed by a directed acyclic graph (DAG), where each vertex $j$ represents a variable $X_j$, with edges going from one variable to another representing causal relationships. It is assumed that each variable $X_j$ is independent of other variables but for its parents, ${\text{pa}}_j$ in the DAG, thus allowing a compressed representation of the joint probability as
%\textcolor{blue}{should we remove this? or change it?}
\begin{equation}
    P(X_1, X_2, \ldots, X_N) = \prod_{j = 1}^N P(X_j \mid \mathbf{\text{pa}}_j).
\end{equation}
The most common approach to exact inference in Bayesian networks is the variable elimination algorithm\citep{PEARL198829}. Approximate inference algorithms are also often applied. The most common one is the Markov Chain Monte-Carlo (MCMC) algorithm that repeatedly samples from each variable conditioned on the values of its parents. The MCMC algorithm predates Bayesian networks and is often referred to as Gibbs sampling.

In relation to temporal data, the Dynamic Bayesian Networks (DBN) are a well-known extension\citep{dbnsfirst,murphythesis}.
DBNs are defined by two Bayesian networks. The first defines the initial state, and the second is the transition model between $t$ and $t+1$, where nodes in layer $t$ are assumed to be independent. The network can then be unrolled into length $T$ so that each of the time slices for $t \geq 1$ is defined by the transition model.

%\paragraph{Structural Equation Modelling}

\paragraph{Counterfactual Framework}

The counterfactual framework can be used to derive causality. This approach focuses on the question of which input variable needs to change in order to change the output of a model. The counterfactual framework is connected with the calculation of interventions, i.e., assessing the change of output variables after a hypothetical change of an input variable. In counterfactuals, we ask which inputs need to change to observe a change in the output, while in intervention, we change the inputs to see the change in the output. %The statement that the patient would survive if he had been treated with drug B instead of drug A or arrived at the hospital two days earlier is an example of counterfactual. On the contrary, the statement that if we change treatment from A to B, the survival probability grows two times is an example of intervention.
%Counterfactuals and interventions can be modeled with many frameworks and have been studied by philosophers in antiquity. 
The counterfactuals were introduced into Bayesian networks by Pearl \cite{Neuberg_2003}. Nowadays, their usage is broad, and they find usage in explainable machine learning models\citep{explainableml}.
%Counterfactuals are also used in deployment, for example, to explain decisions of AI-based models predicting whether an applicant should be granted a loan \cite{explainableinpraxis}. In many cases, the application of counterfactuals to the real world also has limitations, as it brings additional risks, for example, adversarial \cite{NEURIPS2021_009c434c}. Methodologically, counterfactuals can be found by optimization methods \cite{optimizationcounterfact}, heuristical search \cite{cadex}, nearest sample methods \cite{NICE}, as well as others.

\paragraph{Granger Causality}

The goal of the Granger Causality is to detect a causal effect of a time series on another time series\citep{granger}. The Granger causality measures correlations between the effect series and shifted cause series, thus detecting a lag that represents the time needed for the cause to take shape. The method uses various statistical tests to detect whether adding a cause into a predictive model significantly improves the prediction capabilities of the model.
The original paper used linear regression as the testing predictor\citep{granger}. Further modifications of the original paper followed and included non-linearity\citep{wismuller2021}, learning from multiple time series\citep{grangermultiple}, applications on spectral data, i.e., in the frequency domain\citep{kaminski}, model-free modifications \citep{davis2016sparse}, and nonstationary\citep{nonstationary}. 

%\paragraph{Bayesian Structure Time Series Models}

\paragraph{Instrumental Variables}

Instrumental variables can be used to infer causal effects when we cannot control the experimental setting. Suppose that we want to assess the causal effect of the explanatory variable $E$ on the dependent variable $D$. Normally, we would try to do statistical tests on whether variable $D$ changes when $E$ changes. However, in many applications in medicine, economics, and others, this is not possible, as both $E$ and $D$ can have a common cause and be, therefore, correlated. This introduces bias in many statistical tests. To overcome the issue, we include a third variable, instrument variable $I$, which we can control and which has influence on $D$ only through $E$. Then, we observe changes of $D$ on $I$.
When the applied predictor is linear regression, the predictor is a special case of a linear dynamic system\citep{oivr}. The existence of a hidden state then allows the removal of the correlations stemming from a common, unobserved cause\citep{oivr}. 

%Another extension focuses on situations when there is a threshold that changes the influence of a variable on its effect \cite{caner2004instrumental}.

%Other modifications include application to online learning scenarios when it is infeasible to store the whole dataset. Work \cite{oivr} is an example with no-regret performance guarantees with respect to the case when the whole dataset is available. Another extension focuses on situations when there is a threshold that changes the influence of a variable on its effect \cite{caner2004instrumental}. Such situations are widely applicable to economics, where behaviour changes only when input variables cross a threshold, for example, with interest rates, wages, or pollution.

Instrumental variables are, however, concepts that can be used well beyond linear regression. Non-linear\citep{nonlineariv} and non-smooth\citep{caner2004instrumental} modifications exist.
Sometimes, there is a requirement that instrumental variables might have common cofounders. This multilevel modelling is implemented in the instrumental variable toolkit by \cite{Kim2007}. Similarly, \cite{ebbes2005solving} allows for a latent (hidden) variable.

%Paper \cite{ebbes2005solving} provides a solution named Latent Instrumental Variables, where a discrete latent variable model is utilised. Another method, in \cite{momentsIV}, uses higher-order moments. Such and other methods are being implemented in the contemporary packages for instrumental variables as, for example, the R implementation in REndo \cite{rendo}.
% IV regression special case of LDS ... https://ojs.aaai.org/index.php/AAAI/article/view/10215

\paragraph{Tractable Probabilistic Models}

The tractable probabilistic models (TPMs) are a large group of methods that can be used to model probabilistic distributions compactly in the spirit of neural networks approximating functions. A prime example of TPMs is sum-product networks (SPNs)\citep{spns}, which represent the probability distribution as a DAG, where ``input'' random variables are assigned to leaves. Each non-leaf node corresponds to one of two operations, either sum or product. The weights of the edges are then used to learn the probability distribution. The original paper also proposed an algorithm to learn the structure using backpropagation and expectation maximization\citep{spns}. 
The SPNs are only a subgroup in the broad class of probabilistic circuits\citep{choi2020probabilistic}. The unified formalism allows using different types of nodes besides the sum and product nodes. 
Dynamic versions[ \cite{melibari2016dynamic}, e.g.] are able to work with temporal data.

See also Table \ref{tab:capabilities} in the next section for an overview. 

%While sum nodes represent weighted sum, and product nodes represent multiplications, other, so called gate nodes, can exist. The gate nodes can do, for example, logical operations, thresholding, or softmax operations. Terminal (leaf) nodes represent the input probability distribution in all of the probabilistic circuits. From this terminology stems types of circuits as characteristic circuits \TODO{cite}, probabilistic integral circuits, arithmetic circuits, or quantum circuits.
%TODO SPNs, Probabilistic Circuits, characteristic circuits, strudel (?), probabilistic integral circuits

\section{The Challenge}
As suggested in the Introduction, we would like to learn causal models that are more expressive than many traditional models.
In our view, the expressivity of the causal model entails:

\begin{itemize}
%\item %\textbf{Quantitative} aspects of causality, also in order to simulate from the causal model;
\item \textbf{Non-linear} aspects of causality.
\item \textbf{Hidden states} (latent variables) of an \emph{a priori} unknown dimension.
\item At the same time, one would like to preserve as much \textbf{explainability} as possible, perhaps through targeted reduction\citep{kekic2023targeted}.
\item \textbf{Cycles} in causal relationships.
\item \textbf{Time-series} aspects, such as nonanticipativity and delays: clearly, causal relationships should be established between the cause in the past and the effect in the future, with some delay between the two.
\item \textbf{Mixture-model} aspects: clearly, there are variations between the metabolism in various individuals, perhaps due to genomic differences. One should explore joint problems\citep{niu2025joint}, where multiple causal models are learned without the assignment of individuals to subgroups represented by the causal models given \emph{a priori}. 
\end{itemize}
The ability to simulate from the model entails:
\begin{itemize}
\item \textbf{Quantitative} aspects of causality, in order to simulate from the causal model.
\item \textbf{Time} required to simulate from the model scaling modestly (with the number of random variables and numbers of samples).
\end{itemize}

The ability to learn the model entails: 
\begin{itemize}
\item
 \textbf{Sample complexity}: number of samples required to build the model. Even simple models such as HMM comprise learning Gaussian mixture models, which are known to have high sample complexity. 
\item \textbf{Time complexity}: time required to learn the model. Again, even HMM are cryptographically hard to learn in the setting where one has access to i.i.d. samples of observation sequences\citep{blum1993cryptographic,mahajan2023learning}.
\end{itemize}

\begin{table}
    \centering
    \newcommand{\angl}{90}
    \resizebox{\columnwidth}{!}{%
    \begin{tabular}{|l|ccccccc|c|ccc|}
       \hline
        Tool & \rotatebox[]{\angl}{Quantitative} & \rotatebox[]{\angl}{Non-linear} & \rotatebox[]{\angl}{Hidden st.} & \rotatebox[]{\angl}{Cycles} & \rotatebox[]{\angl}{Temporal} & \rotatebox[]{\angl}{Mixture-models} & \rotatebox[]{\angl}{Multiple trajectories} & \rotatebox[]{\angl}{Structure learning in $\mathcal{P}$} & \rotatebox[]{\angl}{Likelihood calculation in $\mathcal{P}$} & \rotatebox[]{\angl}{Marginalization in $\mathcal{P}$} & \rotatebox[]{\angl}{Simulation from the model} \\
       \toprule
        %Pearl's Causal Framework &  &  &  &  &  &  \\  % nor really meant to be there - it discusses counterfactuals, interventions, etc. in BNs and SEM
        Causal Bayesian networks & \yes  & \yes & \yes & \no\footnote{Under some conditions the Bayesian networks can be used to model cyclic relationships. For example, in the case of time series, Dynamic Bayesian Networks can be used to model cyclic relationships between variables at different times. For example $X_{t+1}$ may depend on $Y_{t+1}$, which depend on $X_t$.} & \yes & \yes & \yes & \no & \yes & \no & \yes \\  % quant = continuous variables, latent as variable not connected to any observable, temporal = DBNs, mixtures = hidden variable as condition which model to sample from, for example
        Structural Equation Modeling & \yes & \yes & \yes & \yes & \yes & \yes & \yes & \no & - & - & - \\ 
        % quant=of course, we are formulating equations, hidden states should be one feature of SEM, non-linear of course https://books.google.cz/books?id=VcHeAQAAQBAJ&pg=PA57&redir_esc=y#v=onepage&q&f=false, cyclic = recursive structural equation models
        Counterfactual Framework & \yes & \yes & \yes & \yes\footnote{Depends on the method we use to develop counterfactuals.} & \yes & \yes & \yes & \no\footnote{Depends on the model that is used to build counterfactuals.} & - & - & - \\ 
        % quantitative as we know the amount of change of variable, hidden https://link.springer.com/article/10.1007/s10618-021-00818-9, cycles probably yes - depends on the used toolkit, mixtures are clearly coming from hidden state
        Granger Causality & \yes & \yes & \no & \no & \yes & \no & \yes & \yes\footnote{Relates to the original linear version; for other algorithms, the time complexity might differ.} & -\footnote{The Granger causality is not used directly to answer probability queries} & - & -  \\
        % temporal by def, cycles probably no - we detect time lag and cycle would mean 0 time lag, hidden state no, non-linear with for example kernels (papers: https://www.nature.com/articles/s41598-021-87316-6 ), mixture by lack of reference?
        Bayesian Structural Time Series Models & \yes & \yes & \yes & \yes & \yes & \yes & \yes & \no\footnote{The time complexity may depend on used algorithms and complexity of the model.} & - & - & - \\
        % quantitative - of course, through hidden variables, cyclic relationships - I am not sure ..., temporal by definition, mixture - not primary - latent state works different to BNs, it models growth, fall, etc.,
        Instrumental Variables  & \yes & \yes & \yes & \no & \yes & \yes & \yes & \no & - & - & - \\
        Tractable Probabilistic Models &\yes  & \no & \no & \no & \yes\footnote{{melibari2016dynamic}} & \yes & \yes & \no & (\yes) & (\yes) & (\yes) \\
        % multiple trajectories as multiple instruments, for example selection is non-polynomial, non-linear originally now, but there are methods (aka kernels), hidden no, but can be incorporated, quantitative - yes, it calculates correlations
%       \midrule 
%        Schoelkopf's Measure-Theoretic & ? & ? & ? & ? & ? & ? & ? & \no\footnote{The time complexity may depend on used algorithms and complexity of the model.} & - & - & - \\
%        Schoelkopf's w/ Latent Variables & ?  & ? & ? & ? & ? & ? & ? & ? & ? & ? & ? \\         
        % Tractable Probabilistic Models & \multicolumn{7}{To be developed     
%        Our approach & \yes  & \yes & \yes & \yes & \yes & \yes & \yes & \no & \yes & \yes & \yes 
        \bottomrule
    \end{tabular}
    }
    \caption{Summary of features of selected methods and frameworks.}
    \label{tab:capabilities}
\end{table}

%\subsection{Discussion of the Supported Features in Causal Learning}

Let us discuss some of these in more detail.\\
\paragraph{Cycles} Standard Bayesian networks do not normally support cycles between the variables. The causal relationships need to form a directed acyclic graph (DAG). %which prohibits situations as $X$ causes $Y$, which causes $Z$, and $Z$ causes $X$ again.
%Likewise, Granger causality does not support cyclic dependencies by definition. The Granger causality aims to find out whether one variable can be helpful in predicting the future of a second variable. 
As a result, we are detecting some time lag, that the second variable correlated with the first variable shifted to the future. To obtain a cyclic relationship, we would need a sequence of positive time lags that sum together to zero, which is not possible. Under some circumstances, we can model cyclical relationships with Dynamic Bayesian networks (DBNs). For each variable, we have its realisations for time $t = 1, 2, \ldots, T$. As a result, DBNs can then be used to model situations, such as the one when $X_t$ causes $Y_{t+1}$, $Y_{t+1}$ causes $Z_{t+2}$, which in turn causes $X_{t+3}$. %Thus, we have a cyclic relationship allowed due to the discrete-time. 
The overall graph is still a DAG, as there cannot be a cycle within a one-time slice, and neither can a variable have an effect on the past.

\paragraph{Hidden state and Mixture-models} %Columns describing the possibility of 
modelling a hidden state in the model and sampling from the mixture of models are tightly connected, as the second can be reduced to the first. Suppose that we want to model a mixture of two distributions. We can build two separate models for each of the distributions. Then, we introduce a hidden state that models a binary decision, whether we sample from the first or the second distribution.

%\paragraph{Counterfactual framework} In the counterfactual framework, we are interested in answering what needs to be different in independent variables so that the value of a dependent variable changes. For example, a statement that if the salary of a person were $5{,}000$ USD bigger, he would be eligible for a loan is a counterfactual. Such types of queries can be developed in many models, and counterfactuals can be developed in Bayesian networks, with Instrumental variables, or under Structural Equation modelling. As a result, some of the features may not be available in all scenarios. For example, if the underlying model are Bayesian network, cyclical relationships between variables are not available.

\paragraph{Model learning} When we are interested in the time complexity of model learning, the time requirements differ based on the techniques used. The Bayesian networks do not generally have exact polynomial-time learning. In Granger causality, the complexity of mining causal relationships depends on the algorithms and methods used. In the simplest scenarios, we can base the causal relationships on the F-test, which can be calculated in linear time, assuming that the cumulative distribution function of the Fisher–Snedecor distribution (F-distribution) is precomputed.

\paragraph{Non-linear dependencies} In many cases, the possibility of having non-linear models is part of extensions of the original methods. A prominent example of such a method is Granger's causality. The original method was developed with linear dependencies between the features. But further extensions were developed to include nonlinearities, for example, \cite{wismuller2021}. In Bayesian networks, the original version considered only propositional variables\citep{pearl1985bayesian},
but subsequent versions \cite[e.g.]{continuousbayes} considered also continuous variables and non-linear dependencies.

%(i.e., true/false variables) in the network. Those can be easily generalised to discrete variables and further with continuous variables and non-linear dependencies, for example, \cite{continuousbayes}.

%\paragraph{Quantitative aspects} The same as in the case of non-linear dependencies holds with the quantitative aspects. We are often interested in working with continuous variables, which was not supported in the original Bayesian networks \cite{pearl1985bayesian}. Modern algorithms can deal with continuous variables, which is the necessary condition for any non-linear dependencies.

\section{The Benchmark}
\label{sec:benchmark}
\begin{figure*}%[htbp]
  \centering
  \includegraphics[width=\textwidth, clip]{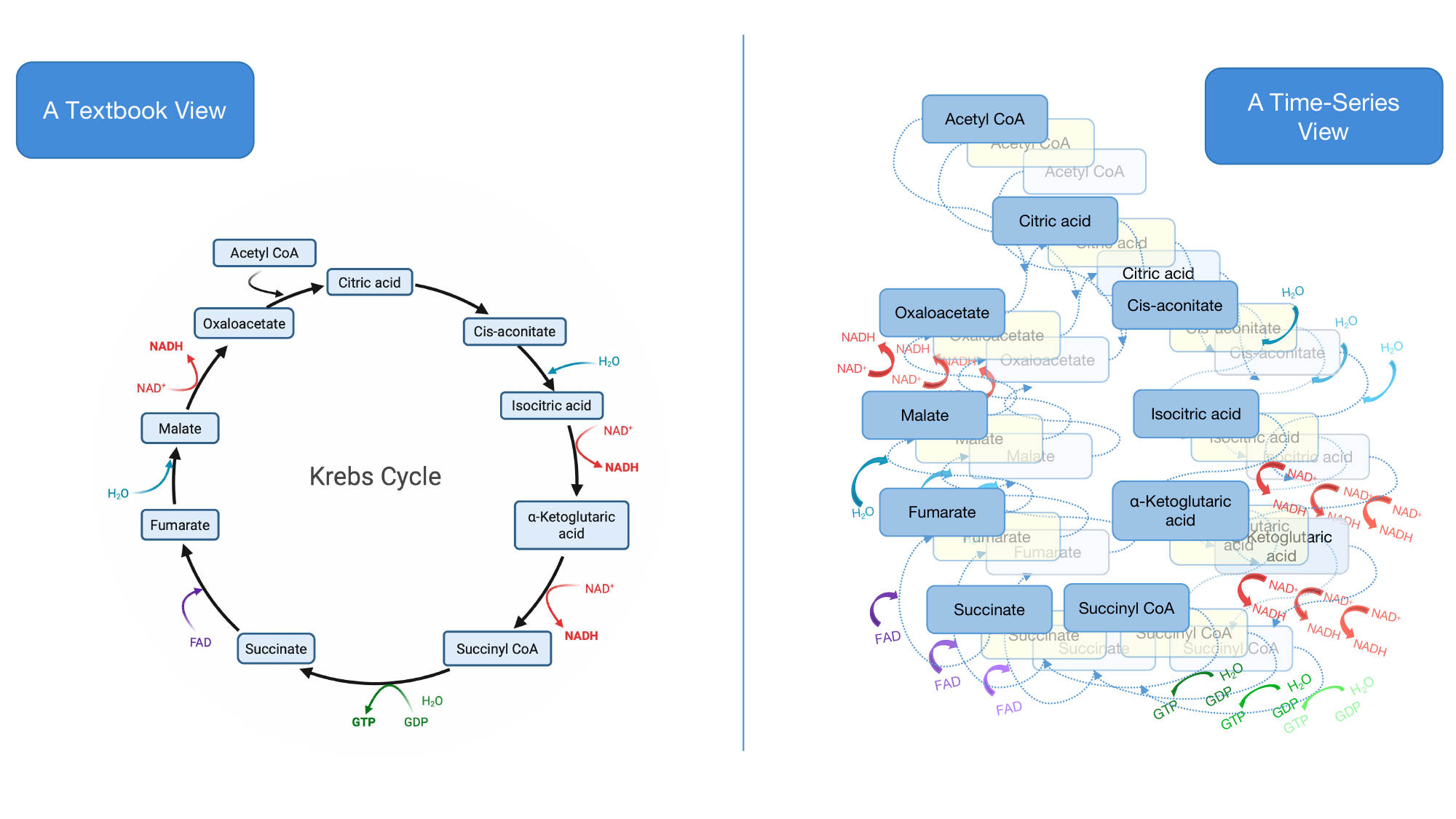}
  \caption{\textbf{Two views of Krebs cycle.}Left: The textbook view, where the nodes representing reactants form a cycle. Right: A time-expanded graph, where nodes represent concentrations of a reactant at one point in time. Nodes corresponding to one reactant could be seen as a time series, but the graph is acyclic in the time-series view.}
  \label{fig:krebs}
\end{figure*}

%As the causal learning community matures, one would like to learn models that are more expressive (see above) and to learn them from datasets that go beyond toy examples. 
In this paper, we present a simulated dataset based on the Krebs cycle. The Krebs cycle, also known as the citric acid cycle, is one of the fundamental pathways of biochemistry. The cycle, as illustrated in Figure \ref{fig:krebs}, presents a natural example of time series that can be used to infer causal relationships between 
concentrations of the reactants.
%The cycle, as illustrated in Figure \ref{fig:krebs}, allows organisms that breathe to convert the energy stored in food to a key energy source (ATP) in muscle cells, for example.

\subsection{The Interventions}
Causal learning requires interventions. Here, we increase the concentration of one reactant and study how its propagations work through the reaction network. In the Krebs cycle, such intervention can be modeled by increasing the concentration of one of the reactants and studying how it propagates through the reaction network. In the natural setting, it is hard to distinguish correlations from causal effects; however, with one reactant artificially increased, an increase in the second reactant means that it is an effect of the first reactant. As this information propagates further, the information about the intervention slowly vanishes, allowing the system to stabilise. In the presented benchmark, this scenario is targeted at in half of the datasets. In krebs1R, only the concentration of one of the reactants is increased. krebs3R presents a more challenging scenario, with three reactants with increased concentrations. Lastly, krebs3 includes normalisation, which is challenging for some of the linear models, but matches situations where we can measure only relative concentrations.

\subsection{The Data}

Depending on the modelling of the time series, each of the reactions can be represented by one or more causal relationships. 
Our benchmark is based on a simulator in the GitHub repository at \cite{github-nagro}. The simulator creates a virtual box with Krebs cycle particles. The particles move inside the box, following the Boltzmann distribution. Once particles get close to each other, a pre-defined list of reactions is scanned to determine whether a reaction occurs, and if so, reactants are replaced with a product. The simulation continues, and concentrations of the particles are noted as time series. As a result, the time series contains noise (caused by the random location of particles), which is added to the locally linear behaviour of the system.

In this way, we have generated four datasets, consisting of a time series with $5$ to $5000$ time steps and $16$ features for the reactants, including $10$ in the main cycle and $6$ additional ones (incl. GTP, H2O, FAD, NAD, GDP, NADH), as depicted in Figure~\ref{fig:krebs}. . Each of the following datasets is based on simulating approximately $2500$ molecules in the bounding box:
\\
\\
\\
\begin{itemize}
    \item[\emph{KrebsN}] contains $100$ series with normally distributed prior distributions and absolute concentrations.
    \item[\emph{Krebs3}] contains  $120$ series with relative concentrations, where for each triplet of the $10$ main cycle reactants, we used uniform priors, and the remaining $7$ particles were set to zero. Such a distribution is motivated by allowing the tested approaches to trace how the higher concentration of the three selected compounds move forward in the cycle.
    \item[\emph{KrebsL}] focuses on learning from a few long time series. In this case, we have $10$ series with $5000$ time steps. We use 
    \item[\emph{KrebsS}] considers  $10000$ time series with only $5$ time steps each, a complementary scenario to \emph{KrebsL}.
\end{itemize}
The datasets are summarized in Table \ref{tab:krebsdata}, showing the dimensions of the time series, the number of molecules used in the simulation, as well as other important features of the data.

\begin{table}
    \centering
    \resizebox{\columnwidth}{!}{%
    \begin{tabular}{l|ccccc}
        \hline
        Dataset & N. features & Lenght & N. series & Initialisation & Concentrations \\
        \hline
        KrebsN & $16$ & $500$ & $100$ & Normal distribution & Absolute \\
        Krebs3 & $16$ & $500$ & $120$ & Excitation of three & Relative \\
        KrebsL & $16$ & $5000$ & $10$ & Normal distribution & Absolute \\
        KrebsS & $16$ & $5$ & $10000$ & Normal distribution & Absolute \\
        \hline
    \end{tabular}
    }
    \caption{\textbf{Summary of the datasets in the Krebs cycle.}}
    \label{tab:krebsdata}
\end{table}

\subsection{Evaluation Criteria}
\begin{figure*}
    \centering
     \begin{subfigure}[b]{0.3\textwidth}
         \centering
         \includegraphics[width=1\textwidth, trim={1cm 0 3cm 1cm}, clip]{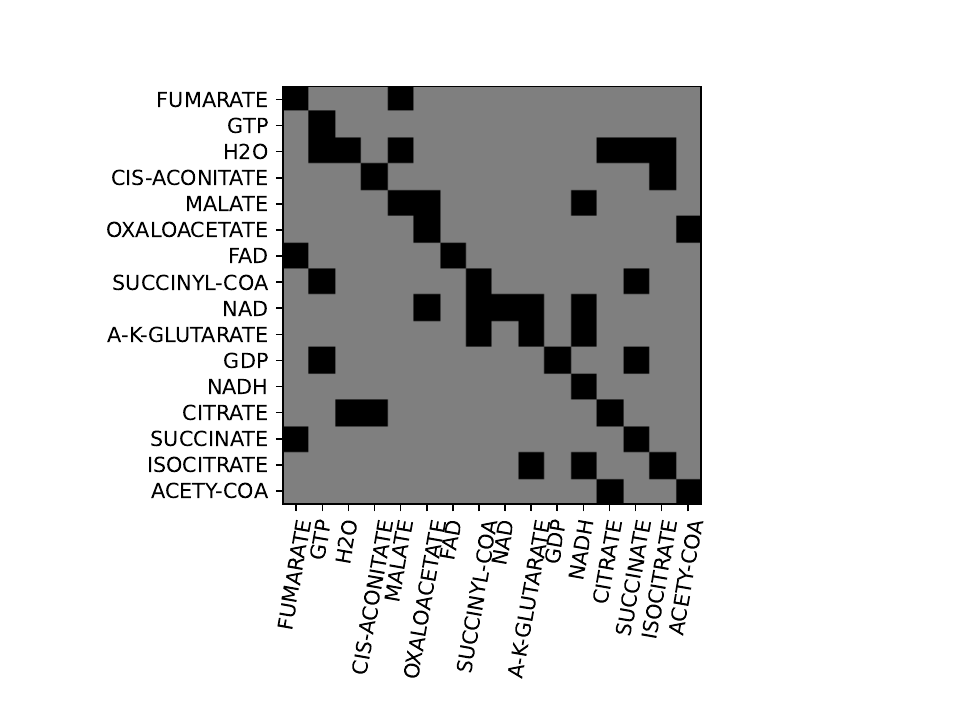}
         \caption{Ground-truth}
         \label{fig:adjacencyGT}
     \end{subfigure}
     \hfill
     \begin{subfigure}[b]{0.3\textwidth}
         \centering
         \includegraphics[width=1\textwidth, trim={1cm 0 3cm 1cm}, clip]{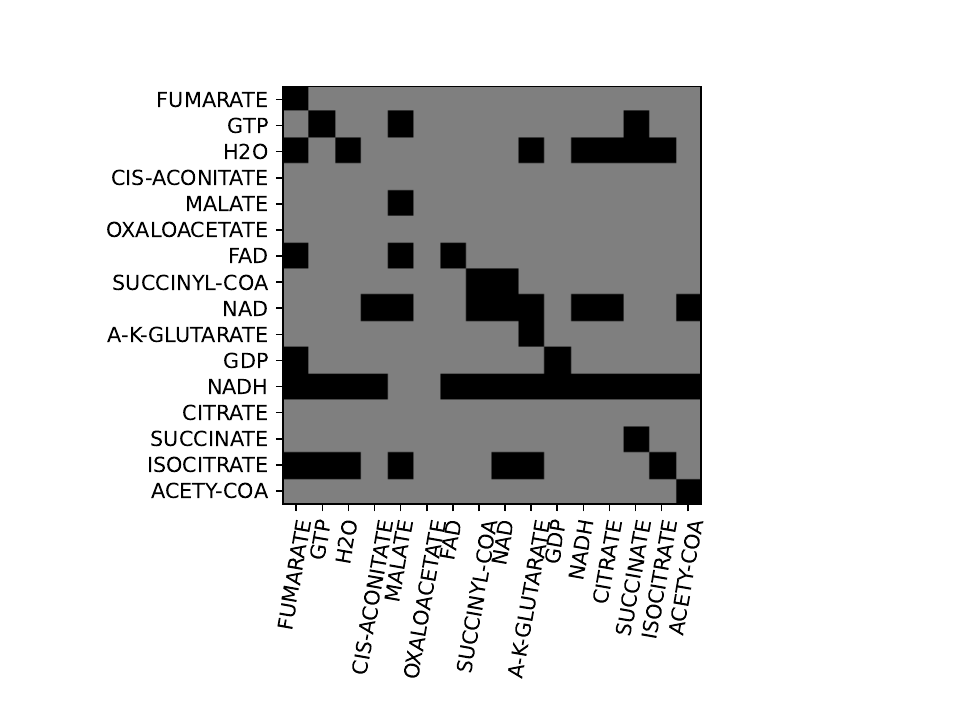}
         \caption{DyNoTears on \emph{KrebsN}}
         \label{fig:adjacencyDNTN}
     \end{subfigure}
     \hfill
     \begin{subfigure}[b]{0.3\textwidth}
         \centering
         \includegraphics[width=1\textwidth, trim={1cm 0 3cm 1cm}, clip]{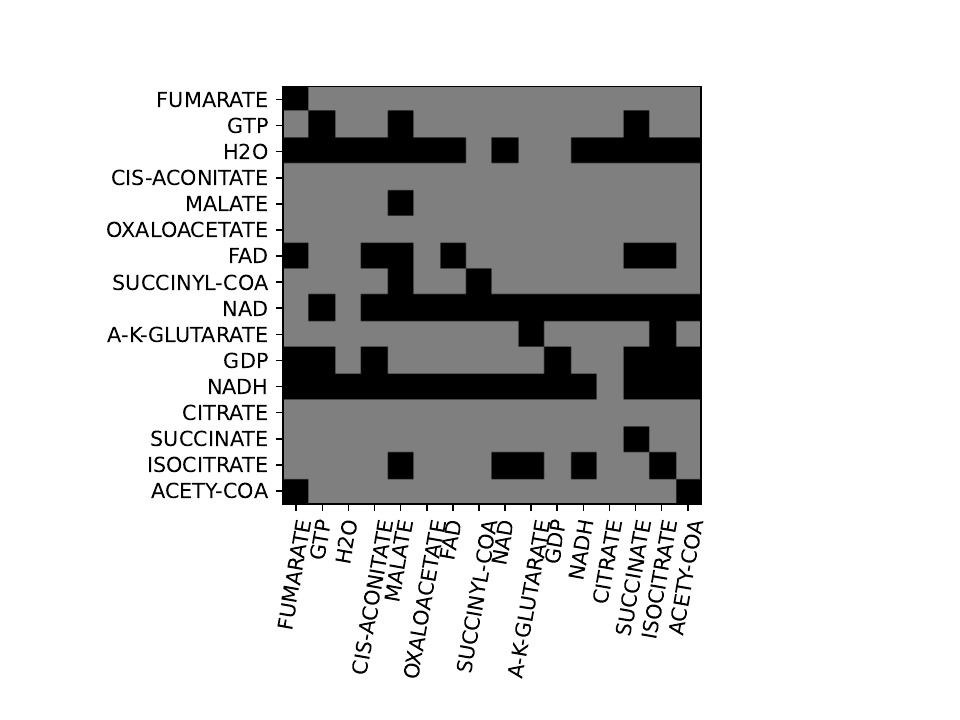}
         \caption{DyNoTears on \emph{Krebs3}}
         \label{fig:adjacencyDNT3}
     \end{subfigure}
     
    \caption{\textbf{An illustration of the adjacency matrix produced by various methods and the ground truth matrix representing the set of reactions.} Black squares represent $1$ an edge in the adjacency matrix, grey $0$.}
    \label{fig:adjacency}
\end{figure*}

% PR:Serene's text?
% 09/13/2024 grrh, it makes no sense to try to do cross-validation if there is GT matrix ... commenting out
%The performance of supervised machine learning methods is compared based on K-fold cross-validation and F-Beta Score (or F1-score). K-Fold Cross-Validation is a technique used to assess the performance and generalisability of a machine learning model. It involves dividing the dataset into $k$ equally sized folds, training the model on $k-1$ folds, and validating it on the remaining fold. This process is repeated $k$ times, with each fold serving as the validation set once. The final performance metric is the average of the metrics obtained in each iteration.

For comparison, the dataset includes the ground-truth causal matrix as defined by the equations. The diagonal in Fig. \ref{fig:adjacencyGT} indicates that the presence of a substance at time $t$ implies the presence of the same substance at time $t+1$. A single run of an algorithm produces a causal matrix that can be compared to the ground truth one. 

We propose that the main measure of the quality of the causal matrix be the Structural Hamming Distance(SHD) and Structural Intervention Distance(SID). SHD measures the number of edges that need to be added and the number of edges that need to be removed to convert the predicted causal graph into the ground-truth causal graph. While the addition or removal of an edge is penalised by $1$, change of the orientation is penalised only by $0.5$. The Structural Intervention Distance (SID)~\citep{peters2015structural} is a metric used to quantify the discrepancy between two causal graphs $\mathcal{G}_{Groundtruth}$ and $\hat{\mathcal{G}}_{Estimated}$,in terms of their implied interventional distributions. Formally, SID counts the number of pairs of variables $(i,j)$ for which the interventional distributions $\mathbb{P}(X_j \mid do(X_i = x_i))$ differ between $\mathcal{G}$ and $\hat{\mathcal{G}}$.
\begin{equation}
\mathrm{SID}(\mathcal{G}, \hat{\mathcal{G}}) = \# \left\{ (i, j) \in \{1, \dots, d\}^2 \mid \mathbb{P}_{\mathcal{G}}(X_j \mid do(X_i)) \neq \mathbb{P}_{\hat{\mathcal{G}}}(X_j \mid do(X_i)) \right\}.
\end{equation}
SID is zero if and only if all pairwise interventional distributions implied by $\hat{\mathcal{G}}$ match those of the true graph $\mathcal{G}$, regardless of the parameterization. Compared to structural Hamming distance, SID focuses on the correctness of interventional implications, making it a more semantically meaningful metric for evaluating causal discovery methods.

$F1$-score, which is the harmonic mean of the precision and recall measures. Let TPR, TNR, FPR, FNR be the true/false positive/negative measures as in a classification task. Then, the False Discovery Rate (FDR), which quantifies the proportion of false positives among all predicted positive instances and $F1$-score is defined 
\begin{align}
    F_1 &=2 \cdot \frac{\textrm{Precision} \cdot \textrm{Recall}}{\textrm{Precision}+ \textrm{Recall}},\\
\textrm{Precision} &=\frac{TPR}{TPR+FPR},\\
\textrm{Recall} &= \frac{TPR}{TPR+FNR},\\
\mathrm{FDR} &= 1 - \text{Precision}.
\end{align}

The $F1$-score can be easily extended to the case where the predicted causal matrix is stochastic. In that case, for example, an edge predicted with weight $0.3$ when there is no ground-truth edge, contributes $0.3$ to false-positive and $0.7$ to true-negative.

%\textcolor{red}{cross-validation F1? mean +- standard deviation time mean +- }

To assess the stability of the method, we recommend to average the results over at least $10$ runs of the method, whenever the tested method is randomized. The standard mean should then be calculated. In the case of deterministic methods, the stability of the $F1$-score cannot be evaluated by simple repeated evaluations followed by standard deviation calculation. Therefore, we recommend using an approach similar to cross-validation to show the stability of the results. In each evaluation, instead of plain restart, we can keep $10\,\%$ of the dataset aside to randomize data instead of the method. As a result, by doing repeated evaluations, it is possible to obtain the results' standard deviations and confidence intervals.
\subsection{Numerical Comparison}

\iffalse
\begin{figure}
    \centering
     \begin{subfigure}[b]{0.45\textwidth}
         \centering
         \includegraphics[width=\textwidth]{groundtruth.pdf}
        \caption{Ground-truth}
        \label{fig:adjacencyGT}
     \end{subfigure}
     \hfill
     \begin{subfigure}[b]{0.45\textwidth}
         \centering
         \includegraphics[width=\textwidth]{series_dynotears.pdf}
         \caption{DyNoTears on \emph{KrebsN}}
        \label{fig:adjacencyDNTN}
     \end{subfigure}
    \caption{\textbf{An illustration of the F1-score of various methods on the Krebs dataset.} Please note }
    \label{fig:f1}
    \hfill
     \begin{subfigure}[b]{0.45\textwidth}
         \centering
         \includegraphics[width=\textwidth]{threes_dynotears.pdf}
        \caption{DyNoTears on \emph{Krebs3}}
        \label{fig:adjacencyDNT3}
        
    \end{subfigure}
    \caption{\textbf{Comparison of learned graphs.} Visualisation of ground-truth and learned adjacency structures using DyNoTears on different subsets of the \emph{Krebs} dataset.}
    \label{fig:adjacencyAll}
\end{figure}
\fi

Then, there is a set of two related methods, ExDAG \citep{exdag}, ExDBN \citep{exdbn}, and ExMAG \citep{exmag}. The methods fit a linear structural equation model to the data. The acyclicity constraints are applied in a lazy manner - whenever a cycle is created in the proposed graph, a new constraint is added to the program, until a DAG is found. ExDAG focuses on learning the data with no interslice dependencies, ExDBN extends the model to the Dynamic Bayesian Networks. Those two models were further extended to the ExMAG method \citep{exmag}, where the goal is to learn a Maximally Ancestral Graph.

The last method in the comparison is DyNoTears \citep{dynotears}, a state-of-the-art method for causal discovery, which was implemented in the CausalNex \citep{causalnex} package. DyNoTears is provided with information that forbids edges within the same time slice, and the regularization parameter $\lambda$ is selected from the list $10^{-6}, 10^{-5}, \ldots, 10^6$, so that the maximum F1-score is reached. In addition to the F1 score, we also measured the time needed for structure learning.
Figure~\ref{fig:adjacency} presents the adjacency matrices inferred by DyNoTears, along with the ground truth. We can see that, as the F1-score is low, both datasets are challenging for causal discovery.
%As evident from the figure, the overall structure recovered by the methods shows limited agreement with the ground truth, highlighting the inherent difficulty of both \textit{KrebsN} and \textit{Krebs3} datasets for causal structure learning, which is also reflected in the low F1-scores.

\begin{figure*}
    \centering
     \begin{subfigure}[b]{0.45\textwidth}
         \centering
         \includegraphics[width=1\textwidth]{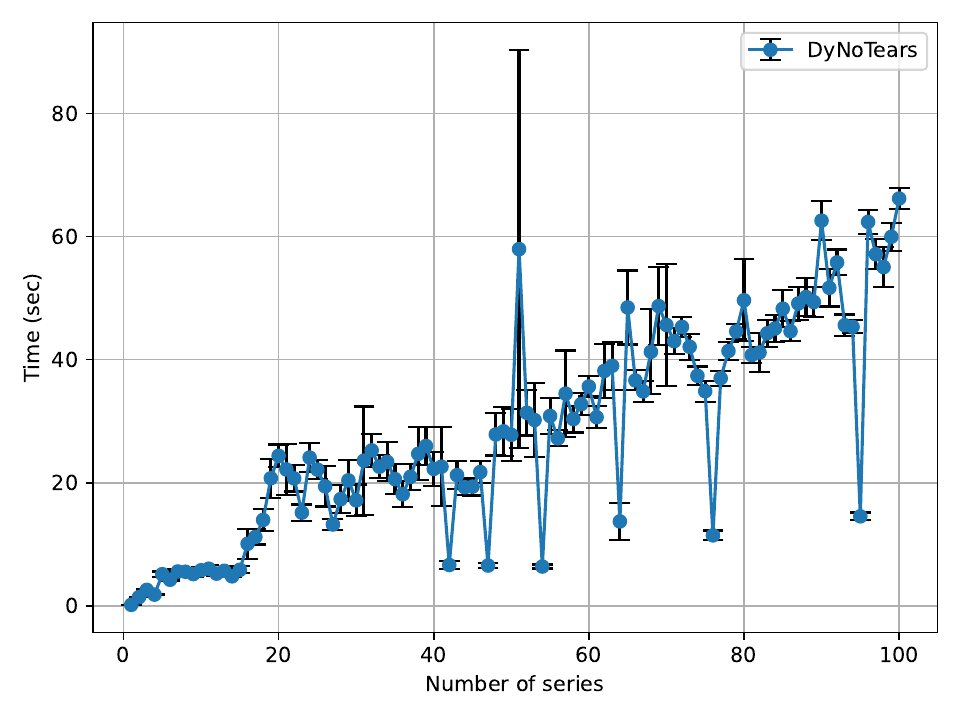}
         \caption{\emph{KrebsN}}
         \label{fig:timeN}
     \end{subfigure}
     \hfill
     \begin{subfigure}[b]{0.45\textwidth}
         \centering
         \includegraphics[width=1\textwidth]{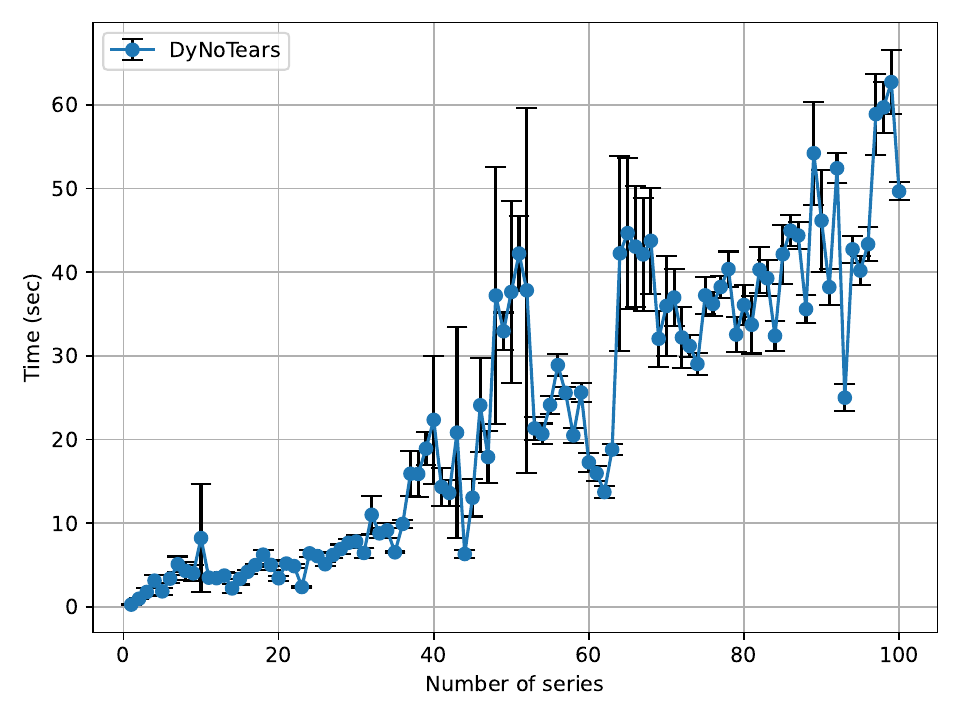}
         \caption{\emph{Krebs3}}
         \label{fig:time3}
     \end{subfigure}
    \caption{\textbf{An illustration of the time required by DyNoTears on the Krebs dataset.} The error bars show the standard deviation of the measurements calculated from 10 repetitions.}
    \label{fig:time}
\end{figure*}

Figure~\ref{fig:f1} illustrates the evolution of the F1-score as a function of the number of time series used in the evaluation. A consistent improvement in performance is observed with an increasing number of time series, suggesting that more data enhances the reliability of the learned causal structure. Similarly, Figure~\ref{fig:time} displays the computational time requirements of the methods across varying dataset sizes. The time complexity increases with the number of time series, but exhibits variability due to the underlying structure and implementation nuances. The error bars indicate standard deviations obtained from 10 repeated runs, providing insights into the stability of each method under varying input conditions.
%Figure \ref{fig:adjacency} shows the adjacency matrix for various methods. We can see that, as the F1-score is low, both datasets are challenging for causal discovery. 
%Figure \ref{fig:f1} shows how the F1-score improves with the number of time series included in the evaluation. Similarly, Fig. \ref{fig:time} shows how the time requirements of the methods change with the number of time series.
From the results, we can see that the dataset is a major challenge for state-of-the-art identification methods, considering their F1-score is close to $0.5$. Therefore, there is room for methods to improve the results further.

\begin{figure}
    \centering
     \begin{subfigure}[b]{0.5\textwidth}
         \centering
         \includegraphics[width=\textwidth]{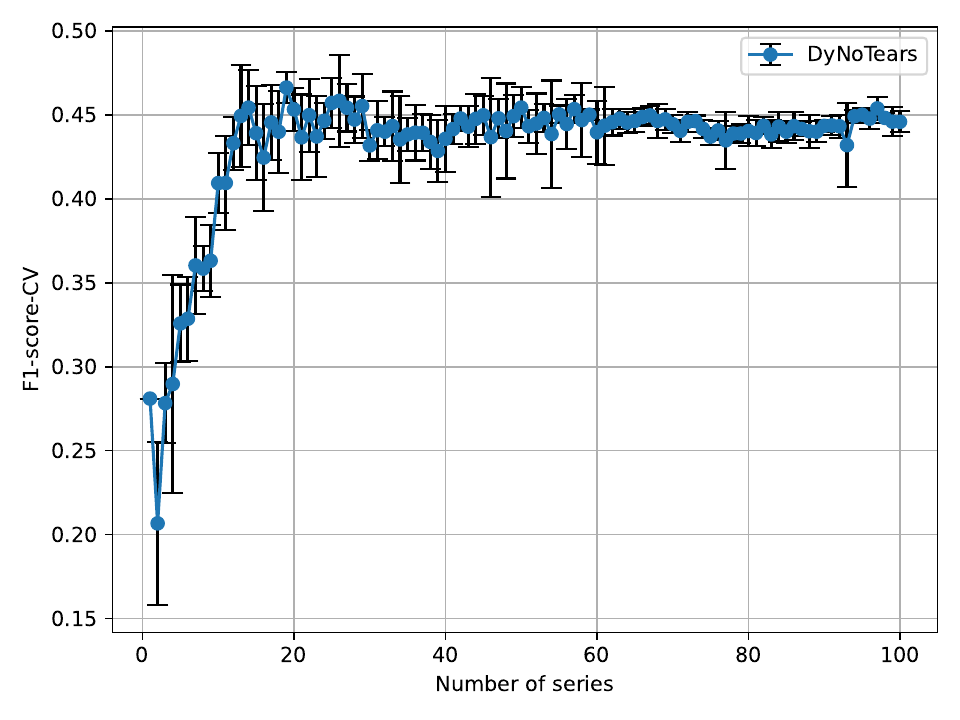}
         \caption{\emph{KrebsN}}
         \label{fig:f1N}
     \end{subfigure}
     \hfill
     \begin{subfigure}[b]{0.5\textwidth}
         \centering
         \includegraphics[width=\textwidth]{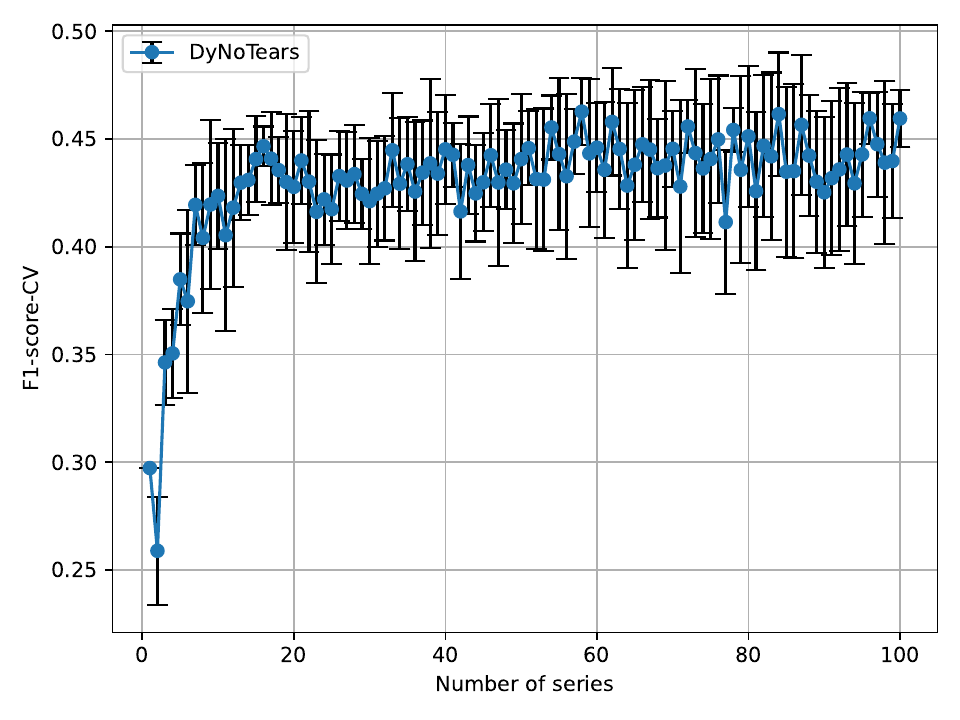}
         \caption{\emph{Krebs3}}
         \label{fig:F13}
     \end{subfigure}
    \caption{\textbf{An illustration of the F1-score of various methods on the Krebs dataset.} Please note that the implementation of DyNoTears in CausalNex is deterministic, thus providing the same result each time. To calculate the error bars, randomly selected $10\,\%$ of the data were put aside, and then results were averaged over $10$ repeats of this procedure.}
    \label{fig:f1}
\end{figure}

To illustrate the dataset, we additionally include results of several representative causal discovery methods that were implemented in the \texttt{GCastle} package.
These methods span a diverse set of causal inference paradigms, including:

\begin{itemize}
    \item \textbf{Constraint-based methods}, such as the PC and FCI algorithms~\citep{spirtes2000causation, zhang2008causal}, which rely on conditional independence testing;
    \item \textbf{Score-based approaches}, including GES and GIES~\citep{chickering2002optimal, hauser2012characterization}, that search for the best causal graph according to a predefined scoring criterion;
    \item \textbf{Functional causal models}, such as LiNGAM~\citep{shimizu2006linear}, which assume linear non-Gaussian causal mechanisms;
    \item \textbf{Gradient-based and deep learning methods}, including NOTEARS\citep{zheng2018dags}, DAG-GNN\citep{yu2019dag}, and GraN-DAG\citep{lachapelle2019gradient}, which formulate structure learning as a continuous optimization problem over the space of acyclic graphs.
\end{itemize}

\begin{figure}
    \centering
    \includegraphics[width=0.46\textwidth]{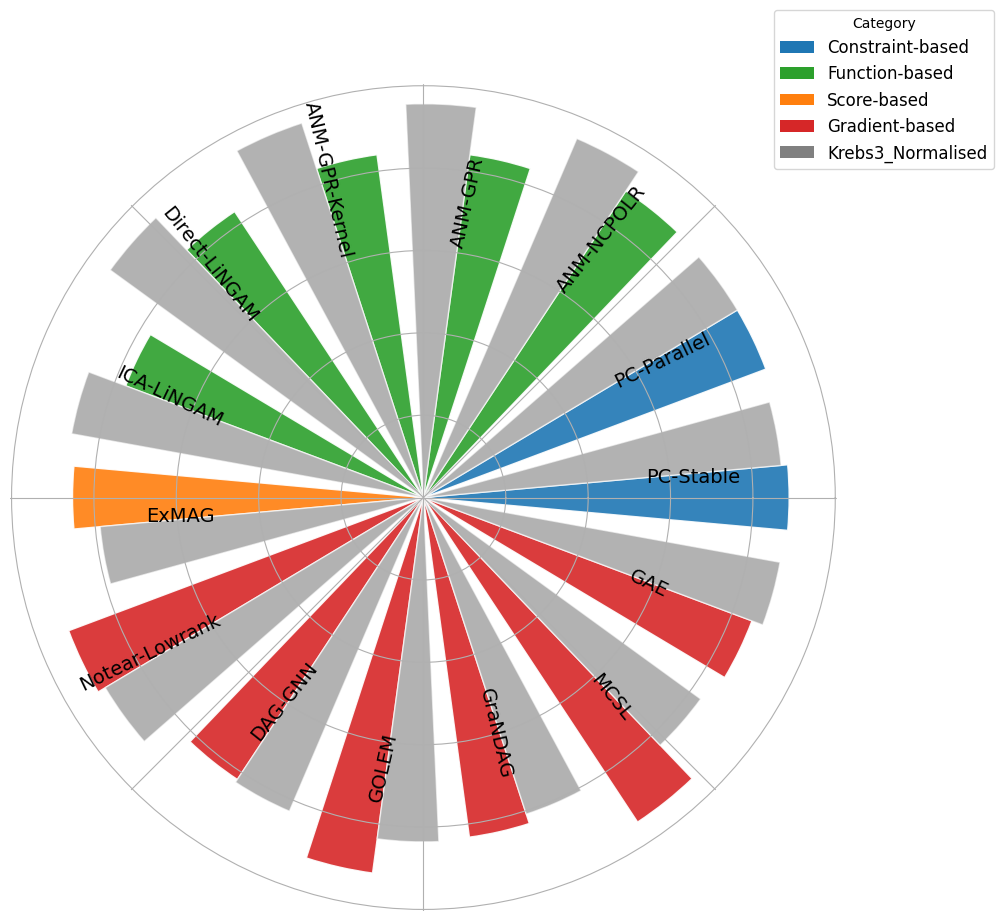}
     \caption{\textbf{SID Scores of Various Algorithms Grouped by Category for Krebs3 and KrebsN datasets.} The performance of each method is represented with color-coded sections in a radial bar plot. The grey overlay indicates the performance on the normalised dataset.}
     \label{fig:Circle_Barplot}
\end{figure}

This variety enables a comprehensive comparison of different causal discovery algorithms across a range of assumptions and data characteristics. To systematically compare the performance of causal discovery algorithms on biochemical pathway data, we evaluated 14 representative methods across four major methodological categories in Figure \ref{fig:Circle_Barplot}. Performance was evaluated on both the original and normalised versions of the \textit{Krebs3} dataset. This comprehensive benchmarking highlights variability in method robustness to data normalisation and facilitates category-level insights into algorithmic behaviour.

\begin{figure*}
    \centering
    \includegraphics[width=0.85\textwidth]{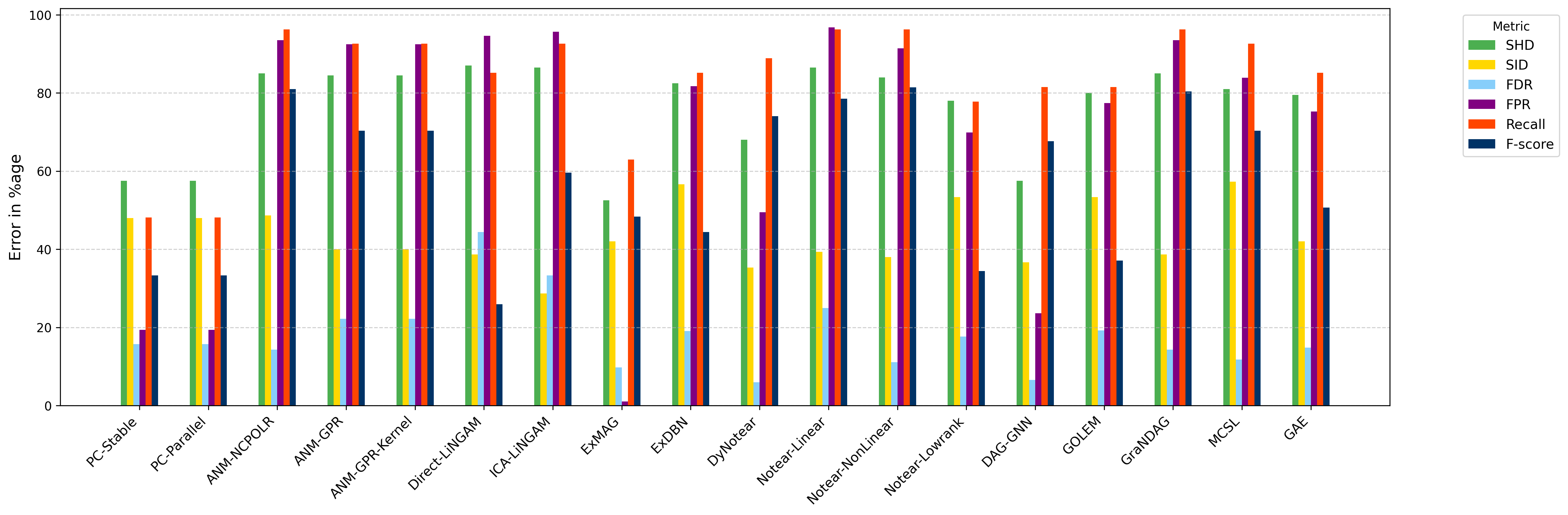}
     \caption{\textbf{Comparison of Various Algorithms by percentage of error in Performance Metrics for $Krebs3$ Dataset.}}
     \label{fig:Barplot}
\end{figure*}

For the \textit{Krebs3} dataset, Figure~\ref{fig:Barplot} illustrates the comparison of different representative causal discovery algorithms using the percentage error across various performance measures. 

In order to calculate \% error, we calculate \% age of absolute difference between the computed value of performance measure and true value to get the percentage error, and is given by the following formula:

\begin{equation}
\text{Percentage Error} = \left| \frac{\text{Computed Value} - \text{True Value}}{\text{True Value}} \right| \times 100.
\label{eq:percentage_error}
\end{equation}

For metrics such as Recall, FDR and FPR, the true value is 1. In the case of SHD and SID, the true value is set to 200 and 150, separately. For F1-score the true values are the respective baseline errors obtained using the ground-truth graph, which are 0.3.

As observed in Figure~\ref{fig:Barplot}, the \texttt{PC} and \texttt{ExMAG} algorithm consistently outperforms all other methods across all evaluation metrics, achieving the lowest percentage errors. This demonstrates its strong performance in both structural and predictive accuracy on large-scale simulated datasets. In contrast, methods like Notear-Linear and DyNotear yield higher error rates, particularly in SHD and F1-score, reflecting challenges in accurately learning causal graphs in this biological context. Overall, the results highlight that method performance can vary substantially depending on the evaluation criterion, emphasizing the importance of using a diverse set of metrics when benchmarking causal discovery algorithms.

\begin{table}
    \centering
    \resizebox{\columnwidth}{!}{%
    \begin{tabular}{c|ccl}
       \hline
       Dataset & $R^2$-sortability & Standard variance & Note \\
       \hline
       krebsN & $0.486$ & $0.008$ & \\
       krebs3 & $0.501$ & $0.011$ & \\
       krebsS & $0.497$ & $0.035$ & (please, see caption) \\
       krebsL & $0.492$ & $0.005$ & \\
    \hline
    \end{tabular}
    }
    \caption{\textbf{The evaluation of $R^2$-sortability for individual time series.} For each of the time series in each dataset, we calculated the $R^2$-sortability using the \texttt{CausalDisco} Python package \citep{reisach1, 
    reisach3}. To obtain the results, the $R^2$-sortability values were then averaged over each of the datasets. Please note that for 11 time series in the \textit{krebsS} dataset, the $R^2$-sortability method did not produce a numeric result. Since this is much less than $1\,\%$ of the dataset (and $R^2$-sortability is bounded by $0$ and $1$), the average won’t be influenced substantially.}
    \label{tab:sortability}
\end{table}

\section{Discussion}

Once the dataset is presented, we are ready to compare it with other existing possibilities and show how it improves upon the other choices in \citep{dynotears, dream4, sachcs}. We will point out the important advantages that the Krebs dataset has over other datasets.

\paragraph{Does Not Assume Any Ground Truth Structural Model.} Instead, our method uses an independent method of simulation from a real-world setting. The dataset is generated by following the chemical reactions in the Krebs cycle. This makes it possible to generate multiple variants (\textit{KrebsN}, \textit{Krebs3}, \textit{KrebsS}, \textit{KrebsS}) consistently. These consist of a time series with $5$ to $5000$ time steps and $16$ features for the reactants, including $10$ in the main cycle and $6$ additional ones (incl. water).

\paragraph{Our Dataset Is Not $R^2$-Sortable.} Our method does not suffer from the $R^2$-sortability issues other synthetic benchmarks suffer from, as explained by \citep{standardizing2} and \citep{reisach3}. Indeed, \cite{standardizing2} argue that there are usually patterns left by the simulation from structural models that are easy to exploit. This can be quantified by the $R^2$-sortability\citep{reisach3}. To illustrate how the Krebs dataset stands compared to the $R^2$-sortability, we implemented a code evaluating the $R^2$-sortability for our dataset, the results of which can be seen in Table \ref{tab:sortability}. Reference\citep{reisach3} then explains that \textit{“0.5 means that ordering the variables by $R^2$ amounts to a random guess of the causal ordering”}, meaning that our dataset is not $R^2$-sortable. Thus, the fact that we do not assume any underlying framework makes our dataset more universal.

\paragraph{The Ground Truth Causal Relationships Are Known.} At the same time, our method comes with widely accepted ground-truth data. The advantage can be seen when compared to datasets such as S\&P100 (stock returns for 100 top US companies), used in DyNoTears paper\citep{dynotears}. S\&P100 is a real-world dataset that suffers from an unclear ground truth causal matrix. Moreover, the authors had to ensure that the data were stationary, as concept drift is likely to happen in stock trading.

A similar situation is connected with the SACHS dataset\citep{sachcs}. This dataset contains single-cell measurements of levels of 11 proteins in immune cells. With ~853 samples, the dataset is of a similar size to ours. However, we cannot be sure what the true causal relationships between the variables representing individual genes are in the case of expression data.

\paragraph{Prospect of Perfect Reconstruction}
At the same time, our method allows for the prospect of perfect reconstruction. Our dataset is much smaller than another commonly used causality dataset, the DREAM dataset\citep{dream4}. This is desirable in connection with the fact that most of the problems in causal learning are NP-hard. Because of that, perfect recovery with many variables is computationally infeasible. \emph{Causal discovery algorithms should be tested on smaller, easy-to-explain datasets first} before proceeding to larger and more complex datasets. The use of larger datasets also brings another reproducibility problem – sampling, often done in an ad hoc, paper-specific fashion –  which is not needed with our data.

\section{Conclusion}
We introduced a synthetic benchmark dataset for causal discovery based on the Krebs cycle. The dataset avoids structural artifacts common in existing benchmarks and includes known ground-truth graphs for evaluation. It supports varied scenarios, including interventions and different time series lengths. Our results show that the dataset poses a meaningful challenge to existing methods and provides a reliable basis for comparing causal inference algorithms.

We publish all source files used to generate the data and the figures in this paper in the GitHub repositories \cite{he2023causal, rysavy2023dynotears}. The repositories also contain numeric results that were generated as input to the plots. 
The generator of the data can be found in another GitHub repository \cite{rysavy2023generator}, including a description of how to generate the benchmarking data. The generator is based on a simulator at \cite{github-nagro}. The dataset is available at \url{https://huggingface.co/datasets/petrrysavy/krebs/tree/main}.

\bibliographystyle{plain}
\bibliography{ref}

%%%%%%%%%%%%%%%%%%%%%%%%%%%%%%%%%%%%%%%%%%%%%%%%%%%%%%%%%%%%

\end{document}